\documentclass[sigconf]{acmart}
\AtBeginDocument{%
  }

\copyrightyear{2026}
\acmYear{2026}
\setcopyright{cc}
\setcctype{by}
\acmConference[ICMI '26]{INTERNATIONAL CONFERENCE ON MULTIMODAL INTERACTION}{October 05--09, 2026}{Napoli, Italy}
\acmBooktitle{INTERNATIONAL CONFERENCE ON MULTIMODAL INTERACTION (ICMI '26), October 05--09, 2026, Napoli, Italy}
\acmDOI{10.1145/3776574.3831133}
\acmISBN{979-8-4007-2318-6/2026/10}

\usepackage{multirow}
\usepackage{array}
\usepackage{booktabs}
\usepackage{caption}
\begin{document}

\title{From Speech to Interaction: Analyzing Multimodal Systems in Cocktail-Party Scenarios}

% \affiliation{%

\author{Thai-Binh Nguyen}
\affiliation{%
  \institution{Karlsruhe Institute of Technology}
  \city{Karlsruhe}
  \country{Germany}
}
\email{thai-binh.nguyen@kit.edu}

\author{Zhaolin Li}
\affiliation{%
  \institution{Karlsruhe Institute of Technology}
  \city{Karlsruhe}
  \country{Germany}
}
\email{zhaolin.li@kit.edu}

\author{Jan Niehues}
\affiliation{%
  \institution{Karlsruhe Institute of Technology}
  \city{Karlsruhe}
  \country{Germany}
}
\email{jan.niehues@kit.edu}

\author{Alexander Waibel}
\affiliation{%
  \institution{Carnegie Mellon University}
  \city{Pittsburgh}
  \country{USA}
}
\email{zhaolin.li@kit.edu}

% \affiliation{%

% \affiliation{%

% \affiliation{%

% \affiliation{%

% \affiliation{%

% \affiliation{%

\renewcommand{\shortauthors}{Thai-Binh Nguyen et al.}

\begin{abstract}
Humans have the remarkable ability to engage in spontaneous informal conversations and selectively attend to individual speakers while filtering out competing speech from nearby conversations. This ``cocktail party'' scenario still presents severe challenges to speech recognition systems. The CHiME-9 MCoRec task provides a testbed where systems must recognize groups of speakers and transcribe each of their conversations from audio-visual input. In this work, we analyze a diverse set of systems, representing different design directions for addressing the cocktail-party scenario, where the best system achieves up to 57\% relative error reduction. We identify three main strategies: (1) explicit or implicit audio-visual target speech separation, (2) improved audio-visual speech recognition for each target speaker, and (3) the use of large language models to group speakers into conversations and enhance conversational consistency. Our analysis shows that these directions address complementary failure modes of the cocktail-party problem, and that high speech overlap alone does not explain performance differences, challenging the common assumption that overlap is the primary source of difficulty in cocktail-party recognition.
\end{abstract}



\begin{teaserfigure}
  \centering
  \includegraphics[width=0.75\textwidth]{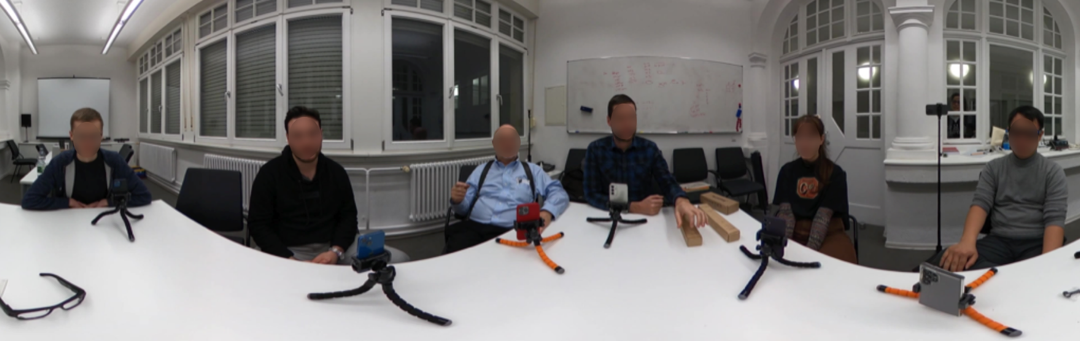}
  \caption{Example recording setup for a multi-party conversation session in a cocktail-party scenario.}
  \Description{An example of a recording session with 6 participants engaged in 3 separate conversations, each consisting of 2 speakers.}
  \label{fig:sample_recording}
\end{teaserfigure}


\maketitle

\vspace{-0.5em}
\section{Introduction}

Recognizing speech in real-world multi-party conversations remains a challenging problem, particularly in so-called cocktail-party scenarios where parallel conversations take place simultaneously in a shared environment. In such settings, systems must handle severe speech overlap, rapid turn-taking, and strong speaker interference, making it difficult to determine not only what is being said, but also who is speaking and with whom they are interacting.

The CHiME-9 Multi-Modal Context-aware Recognition (MCoRec) task provides a realistic benchmark for this problem. Unlike traditional speech recognition tasks that focus on single-speaker or single-conversation settings, MCoRec requires systems to jointly transcribe each speaker’s speech and group speakers into their respective conversations from audio-visual recordings. This introduces additional challenges beyond recognition accuracy, as systems must also resolve speaker ambiguity and conversational structure under highly overlapped and fragmented interactions.

Recent submissions to the MCoRec challenge explore diverse design strategies to address these challenges, including improvements in audio-visual representation learning, explicit target-speaker extraction, and the use of large language models (LLMs) for reasoning about conversational context. While these approaches achieve varying levels of success, it remains unclear which design choices are most effective and what factors truly drive performance differences in cocktail-party recognition.

In this work, we present a comparative analysis of multiple CHiME-9 MCoRec systems, with the goal of understanding how different modeling strategies address the core difficulties of multi-talker recognition. Rather than focusing on individual components in isolation, we analyze systems at the pipeline level and organize them into a unified taxonomy based on how they handle target-speaker recognition, speech separation, and conversation clustering. In addition, we perform condition-aware evaluation to examine system behavior under varying conversational factors, such as speaker activity, number of speakers, and conversational structure.

Our analysis reveals several key findings. First, different system designs address distinct failure modes of the cocktail-party problem, and no single strategy dominates across all conditions. Second, conversation grouping appears to be a comparatively easier problem, as it remains robust to imperfect transcriptions, suggesting that coarse semantic and temporal cues are often sufficient for reliable clustering. Finally, contrary to common assumptions, high speech overlap alone does not fully explain performance differences, suggesting that other factors such as target-speaker representation and conversational modeling play an equally important role. These findings provide a clearer understanding of the challenges in multi-talker recognition and offer guidance for designing more effective multi-modal systems in complex conversational environments.

\vspace{-1em}
\section{Related Work}

% - audio-visual and multimodal datasets 
% - AVSR, target-speaker speech extraction
% - conversation clustering / diarization
% - gaze/face tracks for detecting who speaks with whom.
Audio-visual speech recognition (AVSR) has been extensively studied to improve speech recognition in noisy and multi-speaker environments by leveraging visual cues such as lip movements to complement the acoustic signal \cite{duchnowski94_icslp, suhm1999model, stiefelhagen1999modeling, bub1995knowing, yang1998visual, meier2000towards, duchnowski1995toward, waibe112005chil, yang1999multimodal, gross2000towards}. In overlapping multi-speaker settings, prior work has explored incorporating target-speaker cues or visual speech information directly into the recognition process \cite{chao2016speaker, wu21e_interspeech}. More recent AVSR systems further improve robustness through self-supervised learning, automatic labeling, and stronger cross-modal attention \cite{shi2022robust,10096889,li_unified,rouditchenko2024whisper}. However, most of these methods are developed for single-speaker or controlled conditions and are primarily evaluated based on recognition accuracy, leaving their behavior in realistic cocktail-party scenarios less well understood. 

% Dataset
The renewed interest in audio-visual speech recognition has been accompanied by the availability of large-scale benchmarks. LRS introduced sentence level lip reading in the wild and established a benchmark based on broadcast video \cite{8099850}. LRS3 further scaled this setting with TED talks and became one of the common benchmarks for audio visual speech recognition in unconstrained conditions \cite{afouras2018lrs3}. MuAViC extended this line to multilingual and noise robust learning, supporting both audio visual speech recognition and speech to text translation across multiple languages \cite{anwar23_interspeech}. These benchmarks have progressively increased the realism of AVSR through greater speaker diversity, language coverage, and conversational speech. However, they remain centered on a single visible speaker, moderate acoustic conditions, or prepared speech, and therefore do not capture the concurrent conversational dynamics found in environments such as restaurants, bars, and offices.

Several multimodal corpora have introduced more realistic conversational settings. The CHIL Audiovisual Corpus was one of the first multimodal meeting datasets, introducing synchronized audio, video, and interaction annotations for smart-room meeting and lecture analysis \cite{mostefa2007chil}. The AMI Meeting Corpus provided a benchmark for multimodal meeting analysis with synchronized audio, video, and interaction annotations \cite{ami_2007}. The MISP challenges moved toward home environments and multi participant speech processing \cite{chen22o_interspeech,wang2023multimodal}. The MMCSG dataset further considered multimodal conversations recorded from smart glasses \cite{zmolikova24_chime}. These datasets improve realism, but they still simplify the cocktail party setting as most sessions remain centered on a single shared conversation, and the task mainly focuses on recognition and diarization. As a result, they provide limited support for studying how systems jointly model who speaks, what is said, when speech occurs, and how parallel conversational streams are structured.

Beyond direct audio-visual speech recognition, another line of work uses visual information for target-speaker extraction or speech separation. These methods use a face track, target identity, or cross-modal alignment to recover one speaker from a mixture \cite{ephrat2018looking,gao2021visualvoice, lee2024seeing, pan2025plug}. They provide effective front-end components for noisy and overlapped conditions, but they usually assume a known target speaker and do not address the full ASR task.

Another direction works on audio-visual diarization partially to address the cocktail party problem by studying who spoke when. Recent work has progressed from probabilistic fusion of tracking and sound localization to end-to-end neural activity modeling, late fusion for in-the-wild videos, and multimodal clustering with semantic cues \cite{Gebru_diarization, he22c_interspeech, pan2024late, cheng-etal-2025-integrating}. Although these methods improve diarization in challenging multi-party conditions, they still focus on speaker activity estimation rather than jointly modeling transcription, speaker attribution, and conversation-level interaction structure.

Beyond recognition, conversational clustering is also central to the cocktail-party problem because it helps determine who is speaking to whom. Prior work has used acoustic interaction patterns to infer conversational structure \cite{Boakye08_icassp, dielmann10_interspeech, nguyen2025cocktail}, while visual cues such as gaze and head orientation have been leveraged to model addressee relationships and interaction flow \cite{Vertegaal01_gaze,jovanovic-op-den-akker-2004-towards, chang-etal-2026-multimodal}. More recently, semantic information from dialogue context has been incorporated to infer reply structure and addressee relations in multi-party conversations \cite{le-etal-2019-speaking,gu2021mpc,inoue2025llm}. Despite this progress, these approaches mainly focus on interaction structure alone and do not consider speech content, speaker identity, and conversational flow together in cocktail-party settings.

\vspace{-1em}
\section{Dataset and Tasks}

\subsection{Task Overview}

The MCoRec task~\cite{nguyen2026cocktailpartybenchmarkmultimodaldataset} targets speech recognition in realistic multi-party conversational settings, where multiple speakers may speak simultaneously and form parallel conversations. Unlike conventional ASR tasks that focus solely on transcription, MCoRec requires systems to jointly address three challenges: (i) recognizing speech content, (ii) attributing speech to the correct speaker, and (iii) modeling conversational interactions.

Formally, given a 360$^\circ$ video $V$ and a set of target speakers $S = \{s_1, \dots, s_N\}$, each represented by a sequence of face bounding boxes, the system predicts a set of speaker-dependent transcriptions $\{\hat{Y}_i\}_{i=1}^N$ and a clustering $\hat{C}$ that assigns each speaker to a conversation group. This formulation captures the goal of determining \emph{who speaks what and with whom} in complex multi-speaker environments.

\vspace{-0.5em}
\subsection{Dataset and Recording Setup}

The MCoRec dataset captures natural, unscripted multi-party conversations under realistic conditions. Each session consists of up to eight participants engaged in up to four simultaneous conversations, recorded using a single 360$^\circ$ camera and a single-channel microphone. Figure \ref{fig:sample_recording} shows an example of a recording session with 6 participants engaged in 3 separate conversations, each consisting of 2 speakers.

The recording setup reflects real-world scenarios using consumer devices. A 360$^\circ$ camera is placed at the center of a table, providing a panoramic view of all participants, while the integrated microphone records the entire acoustic scene. The distance between the camera and participants ranges from approximately 0.5\,m to 2\,m. Each session lasts around six minutes and contains spontaneous discussions on everyday topics such as work, hobbies, and personal experiences.

Recordings are collected across diverse indoor environments, including living rooms, meeting rooms, and lecture halls, resulting in varied acoustic and visual conditions. Within each session, speakers are organized into fixed conversation groups, while allowing natural and unconstrained interactions.

A key characteristic of the dataset is the presence of extreme conversational conditions. Due to natural interactions, speech overlap can reach up to 100\%, and conversational turns are highly fragmented. These properties make the task significantly more challenging than conventional multi-speaker benchmarks, requiring systems to handle both acoustic interference and complex interaction dynamics. The MCoRec dataset \footnote{The MCoRec dataset was developed externally by Interactive-AI LLC and is made available for the CHiME Challenge for research purposes. For commercial use, please contact authors. Dataset and
baseline systems can be accessed via \url{https://github.com/MCoRec/mcorec_baseline}.} consists of 150 sessions, divided into training, development, and evaluation splits, with no speaker overlap across splits. A summary of the dataset is shown in Table~\ref{tab:mcorec_summary}.

\vspace{-0.5em}
\subsection{Baseline Pipeline}

The baseline system follows a cascaded architecture consisting of three main components: Active Speaker Detection (ASD), Audio-Visual Speech Recognition (AVSR), and conversation clustering~\cite{nguyen2026cocktailpartybenchmarkmultimodaldataset}.

First, ASD identifies time segments during which each speaker is active, reducing the search space for downstream recognition. These segments are then processed by an AVSR model, which integrates audio and visual cues (e.g., lip movements) to generate speaker-dependent transcriptions under noisy and overlapping conditions. Finally, a clustering module groups speakers into conversations.

The clustering module is based on a temporal overlap assumption: speakers within the same conversation tend to exhibit turn-taking behavior, while speakers from different conversations are more likely to speak simultaneously. Pairwise overlap ratios between speakers are computed and converted into a distance matrix, which is used by an agglomerative hierarchical clustering algorithm to form conversation groups.

\vspace{-0.5em}
\subsection{Evaluation Metrics}

MCoRec evaluates systems using three complementary metrics that capture both transcription accuracy and interaction modeling.

\paragraph{Speaker-dependent WER}
Word error rate (WER) is computed for each speaker based on substitutions, deletions, and insertions, and then averaged across all speakers.

\paragraph{Conversation Clustering F1.}
Clustering performance is evaluated using a pairwise F1 score. For each pair of speakers, the system predicts whether they belong to the same conversation. Precision and recall are computed over all speaker pairs and averaged across sessions.

\paragraph{Joint ASR-Clustering Error.}
The primary evaluation metric combines ASR and clustering into a single score:
\begin{equation}
\text{JointError}(s) = 0.5 \cdot \text{WER}(s) + 0.5 \cdot (1 - F1(s)),
\end{equation}
where $F1(s)$ is computed using a one-vs-rest formulation for each speaker. The final score is averaged over all speakers and sessions.

This metric enforces a balance between transcription accuracy and interaction modeling. In particular, improvements in clustering can significantly impact the overall score, even when WER remains unchanged, highlighting the importance of jointly optimizing both components.

\begin{table}[t]
\centering
\caption{Overview of the MCoRec dataset.}
\vspace{-1 em}
\begin{tabular}{ll}
\toprule
\textbf{Aspect} & \textbf{Description} \\
\midrule
Train / Dev / Eval & 56 / 25 / 67 sessions \\
Duration & 5.6h / 2.5h / 6.9h \\
Speakers per session & 2 -- 8 \\
Conversations per session & 2 -- 4 \\
Recording & 360$^\circ$ video + single-channel audio \\
Extra modality & Ego-view videos (train only) \\
Annotations & Speaker's transcription + groups \\
\bottomrule
\end{tabular}
\label{tab:mcorec_summary}
\vspace{-1.5em}
\end{table}

% \section{Taxonomy and Design Space}

% - cascaded vs end-to-end long-context.
%     + USTC-NERCSLIP cascade (ASD → AVTSE → AVSR → clustering)
%     + BUT: long-context end-to-end target-speaker AVSR

% - explicit AVTSE vs implicit extraction in AVSR
%     + SUSTech explicit AVTSE
%     + Science Tokyo AVSR implicit for target extraction and includes auxiliary AVTSE loss  intermediate representation to joint optimize

% - AV-HuBERT vs Whisper/large-scale ASR vs fusion
%     + USTC-NERCSLIP: Whisper + LLM
%     + MGazE: dual AVSR
%     + BUT: ASR (NVIDIA Parakeet-v2) fuse visual representation from AV-HuBERT to do target-speaker AVSR
%     + NJU: Whisper‑Flamingo 

% - clustering: time/overlap heuristics vs gaze cues vs LLM semantic
%     + USTC-NERCSLIP use zero-shot clustering LLM (F1=1.0)
%     + BUT, NJU: Qwen3 LLM
%     + MGazE: mutual gaze and fuse with overlap-score. (limit to head pose condition)
%     + AUVIS: time/overlap heuristics params search.

\begin{table*}[htbp]
\centering
\caption{Component-level comparison between systems}
\label{tab:detailed_components}
\renewcommand{\arraystretch}{1.15}
\small

\begin{tabular}{
@{}
>{\raggedright\arraybackslash}p{1.8cm}
>{\raggedright\arraybackslash}p{2.4cm}
>{\raggedright\arraybackslash}p{2.8cm}
>{\raggedright\arraybackslash}p{3.4cm}
|
>{\raggedright\arraybackslash}p{2.6cm}
>{\raggedright\arraybackslash}p{3.0cm}
@{}
}
\toprule

\multirow{2}{*}{\textbf{System}} & 
\multicolumn{3}{c|}{\textbf{Target-Speaker Transcription Pipeline}} & 
\multicolumn{2}{c}{\textbf{Conversation Clustering}} \\
\cmidrule(lr){2-4} \cmidrule(l){5-6}

& \textbf{ASD} 
& \textbf{AVTSE (Extraction)} 
& \textbf{AVSR (Recognition)} 
& \textbf{Modality / Cues} 
& \textbf{Algorithm / Model} \\
\midrule

\textbf{Baseline\cite{nguyen2026cocktailpartybenchmarkmultimodaldataset}} & 
Light-ASD & 
- & 
AV-HuBERT CTC/Attention & 
Temporal (Overlap / Turn-taking) & 
Agglomerative Hierarchical Clustering (AHC) \\
\midrule

\textbf{S1 \cite{jiang26_chime}} & 
ConvNeXt + Conformer with Binary Classifier  & 
Ensemble: BRAVEn, ResNet-18 & 
Ensemble: ConvNeXt + Conformer fuse (Whisper/WavLM) + Transformer decoder & 
Semantic (Text) + Timestamps & 
Qwen-2.5 \& DeepSeek R1 (Zero-shot) \\
\midrule

\textbf{S2.1 \cite{song26_chime}} & 
Baseline & 
- & 
AV-HuBERT  & 
Semantic (Text) + Timestamps & 
Qwen3-8B (Deep reasoning) \\
\midrule

\textbf{S2.2 \cite{song26_chime}} & 
Baseline & 
- & 
AV-HuBERT + Qwen3-8B (post correction)  & 
Semantic (Text) + Timestamps & 
Qwen3-8B (Deep reasoning) \\
\midrule

\textbf{S2.3 \cite{song26_chime}} & 
Baseline & 
- & 
Whisper-Flamingo & 
Semantic (Text) + Timestamps & 
Qwen3-8B (Deep reasoning) \\
\midrule

\textbf{S3 \cite{ackermann26_chime}} & 
Baseline (Tuned thresholds) & 
- & 
Baseline & 
Baseline & 
AHC (optimized distance/linkage) \\
\midrule

\textbf{S4 \cite{klement26_chime}} & 
- & 
- & 
Parakeet FastConformer + AV-HuBERT visual features & 
Semantic (Text) + Temporal fallback & 
Qwen3 similarity + AHC (+ fallback) \\
\midrule

\textbf{S5 \cite{hartanto26_chime}} & 
Baseline & 
- (Auxiliary AVTSE loss only) & 
AV-HuBERT (multitask learning) & 
Baseline & 
Baseline \\
\midrule

\textbf{S6.1 \cite{ling26_chime}} & 
Baseline & 
AV-TFGridNet (pseudo-supervised) & 
AV-HuBERT & 
Baseline & 
Baseline \\
\midrule

\textbf{S6.2 \cite{ling26_chime}} & 
Baseline & 
AV-TFGridNet (pseudo-supervised) & 
Muavic-EN & 
Baseline & 
Baseline \\
\midrule

\textbf{S7.1 \cite{li26_chime}} & 
Baseline & 
- & 
AV-HuBERT + Whisper (dual-use fallback policy) & 
Visual (Mutual Gaze) & 
UniGaze + MGaze scoring + AHC \\

\midrule

\textbf{S7.2 \cite{li26_chime}} & 
Baseline & 
- & 
AV-HuBERT + Whisper (dual-use fallback policy) & 
Audio + Visual (Overlap + Gaze) & 
AVFuse (gaze + overlap fusion) + AHC \\

\bottomrule
\end{tabular}
\vspace{-1.5 em}
\end{table*}

\vspace{-0.5em}
\section{Taxonomy and Design Space}

We refer to individual systems using anonymized identifiers\footnote{Each system is described in a separate system paper to be released soon.} (S1--S7) and analyze them through their architectural design choices rather than team identities. Our goal is not only to compare systems by performance, but also to identify the main design axes along which current multimodal cocktail-party systems differ.

We organize the design space around the two components that are directly reflected in the task definition and evaluation: \textit{target-speaker transcription} and \textit{conversation clustering}. This decomposition aligns with the two main sources of error in the benchmark: speaker-dependent word error rate (WER), which captures how well a system transcribes each target speaker, and clustering F1, which captures how well it models interaction structure. Although we discuss these components separately for clarity, they are strongly coupled in practice. Segmentation errors propagate into recognition, recognition quality affects semantic clustering, and failures in either component directly affect the joint evaluation metric.

Table~\ref{tab:detailed_components} summarizes the systems at the component level. Compared with the baseline, the submitted systems span a considerably richer design space. They differ in their target-speaker transcription pipelines, including how speech segments are defined (ASD), whether audio-visual target-speaker extraction (AVTSE) is performed before recognition, and which AVSR backbone is used. They also differ in conversation clustering, depending on whether it is based on temporal overlap, semantic reasoning, or visual interaction cues.

\vspace{-1em}
\subsection{Target-Speaker Transcription Pipelines}

The target-speaker transcription problem in MCoRec can be viewed as a sequence of three subproblems: identifying when a speaker is active, isolating or emphasizing the target speaker under overlap, and mapping the resulting signal to text. Across the systems, we observe three major design dimensions: segmentation strategy, target-speaker extraction strategy and recognition backbone.
% , and training adaptation strategy.
\paragraph{\textbf{Segmentation:} explicit ASD \textnormal{versus} long-context recognition.}
Most systems retain the baseline assumption that recognition should operate on short speaker-dependent segments obtained from an active speaker detection module. The baseline itself uses Light-ASD\cite{ASD} to identify active regions and segment long video into smaller chunks before AVSR. S2, S3, S5, S6, and S7 largely follow this paradigm, though with different levels of dependence on the baseline segmentation. S2, S5, S6 and S7 keep the baseline Light-ASD unchanged and focuses improvements on downstream AVSR. S3 keeps the baseline pipeline but tunes onset/offset and chunking-related parameters to improve segment quality for both recognition and clustering. Two systems (S1 and S4) depart more clearly from this standard setup. S1 replaces the lightweight baseline ASD with a stronger audio-visual detector based on the ConvNeXt encoder \cite{convnet} combined with Conformer\cite{conformer} for frame binary classifiers (this ConvNeXt + Conformer encoder uses both in ASD and AVSR). At the other extreme, S4 avoids explicit short-segment ASD-based recognition and instead performs long-context target-speaker AVSR by concatenating speaker-conditioned visual tracks (lip crop frames) into a continuous stream and decoding substantially longer contexts in a single pass. This shifts part of the burden from segmentation to the recognizer itself.

\begin{table*}[htbp]
\centering
\caption{Datasets used by systems.}
\label{tab:dataset_usage_trained_only}
\renewcommand{\arraystretch}{1.1}
\small
\begin{tabular}{
@{}
>{\raggedright\arraybackslash}p{1.5cm}
>{\raggedright\arraybackslash}p{2.6cm}
>{\raggedright\arraybackslash}p{6.7cm}
>{\raggedright\arraybackslash}p{5.8cm}
@{}
}
\toprule
\textbf{System} & \textbf{Trained component} & \textbf{Datasets used} & \textbf{Development strategy} \\
\midrule

\multirow{3}{*}{\textbf{S1 \cite{jiang26_chime}}} 
& ASD 
& MCoRec, AVA-Speech, MSDWILD, M3SD 
& Multi-dataset ASD training \\

& AVTSE 
& LRS3, VoxCeleb2, AVSpeech, DNS-Noise 
& Synthetic 2--3 speaker mixtures \\

& AVSR 
& LRS2, LRS3, VoxCeleb2, AVSpeech, AVYT, MCoRec. 
& Large-scale AV pretraining, Whisper pseudo-labels \\

\midrule

\multirow{2}{*}{\textbf{S2 \cite{song26_chime}}} 
& AVSR (AV-HuBERT CTC/Attention) 
& LRS2, VoxCeleb2, AVYT, AVYT-mix, MCoRec 
& Pretraining + MCoRec fine-tuning. MCoRec also used to derive overlap templates \\

& AVSR (Whisper-Flamingo) 
& LRS2, VoxCeleb2, AVYT, AVYT-mix, MCoRec 
& End-to-end training with template-driven simulation and short MCoRec adaptation \\

\midrule

\multirow{1}{*}{\textbf{S3 \cite{ackermann26_chime}}} 
& - 
& MCoRec 
& MCoRec dataset to tune hyperparameters \\

\midrule

\multirow{1}{*}{\textbf{S4  \cite{klement26_chime}}} 
& AVSR 
& AVYT, LRS3, AMI, MCoRec 
& Stage-1 simulated pretraining on AVYT+LRS3. Stage-2 on AMI. Fine-tuning on MCoRec (+ simulated AMI) \\

\midrule

\multirow{1}{*}{\textbf{S5  \cite{hartanto26_chime}}} 
& AVASR + auxiliary AVTSE 
& LRS2, AVYT, VoxCeleb2, MCoRec 
& Simulated mixtures from LRS2/AVYT/VoxCeleb2. Final-stage training mixed with MCoRec \\

\midrule

\multirow{1}{*}{\textbf{S6  \cite{ling26_chime}}} 
& AVTSE
& VoxCeleb2-2Mix, VoxCeleb2, MCoRec 
& Pretrained on simulated VoxCeleb2-2Mix and tuned with pseudo labels derived from MCoRec \\

\midrule

\multirow{1}{*}{\textbf{S7  \cite{li26_chime}}} 
& - 
& MCoRec 
& Uses MCoRec dataset to tune hyperparameters of UniGaze, MGaze and AVFuse\\

\bottomrule
\end{tabular}
\vspace{-1 em}
\end{table*}

\paragraph{\textbf{Target-speaker extraction}: explicit extraction \textnormal{versus} implicit conditioning.}
A major axis of variation concerns whether systems explicitly separate or extract target speech before recognition. The baseline does not include a dedicated extraction stage, it relies on visual conditioning inside AVSR to bias recognition toward the visible target speaker. Several submissions continue this implicit strategy like S2, S3, S4, S7. In contrast, S1 and S6 adopt explicit audio-visual target speech extraction (AVTSE). S1 develops four AVTSE variants and uses an ensemble strategy, including BRAVEn-based\cite{braven} semantic-phonetic encoders and a ResNet-18-based dual-tower design\cite{lip-resnet}, to disentangle the target stream before AVSR. S6 follows a two-stage pseudo-supervised paradigm: close-talk recordings are first enhanced to reduce cross-talk, then used as pseudo-labels to fine-tune an AV-TFGridNet\cite{tfgridnet} extraction model that operates on far-field audio plus target visual input. Recognition is subsequently performed on the extracted signal, and ASD is applied after extraction for segmentation. These systems reflect the view that target-speaker isolation should be solved explicitly before transcription.

S5 occupies an intermediate position. It does not perform explicit extraction at inference time, but it introduces an auxiliary AVTSE objective during training. Concretely, it adds a SEANet-style\cite{seanet} reconstruction branch on top of intermediate AVSR representations so that the shared encoder is encouraged to preserve target-speaker acoustic information, while the test-time path remains a standard AVSR decoder.

\paragraph{\textbf{Recognition backbones:}} AV-HuBERT\cite{avhubert, avhubert_cocktail} dominance, but with different extensions.
The most common recognition backbone is AV-HuBERT with CTC/Attention decoding\cite{ctcattn}. This is the baseline model and remains the reference point for most submissions. S2 uses it as one of two main AVSR options. S3 keeps it and improves results mainly through better segmentation and decoding parameters. S5 builds directly on AV-HuBERT and strengthens it via multitask training. S6 evaluates several AVSR backends on top of extracted speech, including baseline AV-HuBERT, Muavic-EN\cite{muavic}. S7 combines baseline AV-HuBERT with a Whisper-based system in a dual-model\cite{dualdecode} inference policy: whenever the AV-HuBERT model produces an empty output, whisper-based replace it by the dual-use hypothesis to potentially fill the gap.

Other systems push beyond the standard AV-HuBERT recipe in more radical ways. S1 adopts a heterogeneous ensemble of AVSR models spanning both self-supervised and large-scale encoder–decoder paradigms. The ensemble includes a ConvNeXt Conformer encoder trained with masked audio-visual pretraining and discrete unit prediction, as well as Whisper-based\cite{whisper} encoder–decoder models augmented with Flamingo-style\cite{flamingo} gated cross-attention for visual conditioning. Additional variants leverage pretrained speech representations such as WavLM to improve acoustic robustness. Some branches further incorporate LLM-conditioned decoding (e.g., Qwen-based refinement) to enhance linguistic coherence and error correction. Final hypotheses are combined via posterior-level fusion and ROVER\cite{rover} system combination to exploit complementarity across models. S2 also explores Whisper-Flamingo\cite{whisper-flamingo} as an alternative to AV-HuBERT, investigating how a large single-speaker ASR backbone can be adapted to the multi-speaker audiovisual condition. S4 takes a notably different route by conditioning a Parakeet FastConformer/TDT\cite{fastconformer, tdt} recognizer on baseline encoder AV-HuBERT's visual features, allowing long-context decoding with a strong audio backbone rather than using a conventional AV-HuBERT encoder-decoder stack.

\vspace{-0.5em}
\subsection{Conversation Clustering Strategies}

The second component is to group speakers into conversation-level interactions. Here the design space is even more diverse than for transcription. We observe three broad families of approaches: temporal-overlap heuristics, semantic clustering based on transcripts, and multimodal interaction modeling using visual cues. The main distinction is the level of abstraction at which the system reasons about interaction: low-level activity patterns, utterance content, or speaker attention cues.

\paragraph{\textbf{Temporal clustering from speech activity:}}
The baseline clustering module derives pairwise speaker similarity from overlap ratios between active speaking regions and then applies agglomerative hierarchical clustering (AHC)\cite{nguyen2026mcorec}. The core assumption is that speakers in the same conversation tend to alternate, whereas speakers from different conversations overlap more often. S3 remains closest to this philosophy, retaining the same basic temporal clustering approach but tuning the distance and linkage hyperparameters to better match the MCoRec development set. S5 and S6 also keep the baseline clustering unchanged. This approach is attractive because it's simple, efficient, and does not depend on linguistic quality. However, it is also tightly constrained by the quality of ASD or post-extraction segmentation. If activity estimates are noisy, temporal clustering becomes unreliable. This dependence is especially important in MCoRec, where overlaps are frequent and utterances are fragmented, so even small segmentation shifts can alter pairwise overlap statistics substantially.

\paragraph{\textbf{Semantic clustering with large language models:}}
A second cluster of systems treats conversation clustering as a semantic reasoning problem rather than a temporal one. S1 uses large language models, including Qwen 2.5\cite{qwen25}(70B) and DeepSeek R1\cite{DeepSeek-R1}(671B), in a multi-stage zero-shot prompting pipeline that selects speaker-to-conversation assignments from transcripts and timestamps. S2 uses Qwen3-8B\cite{yang2025qwen3technicalreport} with deep reasoning mode, explicitly prompting the model to infer conversational grouping from recognized text and utterance timing. S4 uses Qwen3 to derive semantic relations between speakers: it first classifies speakers as topic-bearing or passive based on their transcripts, then estimates pairwise topic similarity scores between active speakers. These scores form a similarity matrix, on which AHC is applied to obtain core conversation groups. Passive speakers are finally assigned to these groups using a temporal overlap-based fallback.

Compared with temporal clustering, these systems reason at the level of discourse coherence rather than local turn-taking. This is potentially much more powerful in MCoRec because simultaneous conversations are often separable by topic even when speaking patterns are ambiguous. However, semantic clustering is only as reliable as the transcripts it receives. Recognition errors, short backchannels, and semantically sparse utterances can all reduce the usefulness of LLM-based grouping, which is why S4 explicitly adds an acoustic speech activity fallback for passive or low-content speakers.

\begin{table*}[htbp]
\centering
\caption{Development and evaluation results. Ranks are assigned within each split based on JointError.}
\label{tab:dev_eval_results}
\renewcommand{\arraystretch}{1.1}
\setlength{\tabcolsep}{5pt}
\normalsize
\begin{tabular}{
@{}
l
c c c c
c c c c
@{}
}
\toprule
\multirow{2}{*}{\textbf{System}} &
\multicolumn{4}{c}{\textbf{DEV}} &
\multicolumn{4}{c}{\textbf{EVAL}} \\
\cmidrule(lr){2-5} \cmidrule(l){6-9}
&
\textbf{Rank} &
\textbf{WER $\downarrow$} &
\textbf{Conv F1 $\uparrow$} &
\textbf{JointError $\downarrow$} &
\textbf{Rank} &
\textbf{WER $\downarrow$} &
\textbf{Conv F1 $\uparrow$} &
\textbf{JointError $\downarrow$} \\
\midrule
S1   & 1  & $\textbf{0.3140} \pm 0.0264$ & $\textbf{1.0000} \pm 0.0000$ & $\textbf{0.1570} \pm 0.0132$ & 1  & $\textbf{0.3018} \pm 0.0208$ & $\textbf{0.9572} \pm 0.0448$ & $\textbf{0.1597} \pm 0.0173$ \\
S2.1 & 3  & $0.4288 \pm 0.0198$ & $0.9778 \pm 0.0333$ & $0.2264 \pm 0.0258$ & 4  & $0.4525 \pm 0.0231$ & $\underline{0.9545} \pm 0.0452$ & $0.2360 \pm 0.0187$ \\
S2.2 & 3  & $0.4288 \pm 0.0197$ & $0.9778 \pm 0.0333$ & $0.2264 \pm 0.0246$ & 3  & $0.4451 \pm 0.0219$ & $\underline{0.9545} \pm 0.0469$ & $0.2323 \pm 0.0181$ \\
S2.3 & 7  & $0.4877 \pm 0.0226$ & $0.9778 \pm 0.0333$ & $0.2559 \pm 0.0258$ & 5  & $0.4952 \pm 0.0270$ & $0.9442 \pm 0.0471$ & $0.2689 \pm 0.0255$ \\
S3   & 8  & $0.4931 \pm 0.0165$ & $0.9028 \pm 0.0691$ & $0.2899 \pm 0.0379$ & 9  & $0.5306 \pm 0.0332$ & $0.8137 \pm 0.0606$ & $0.3639 \pm 0.0365$ \\
S4   & 2  & $\underline{0.3369} \pm 0.0271$ & $0.9667 \pm 0.0433$ & $\underline{0.1804} \pm 0.0255$ & 2  & $\underline{0.3034} \pm 0.0219$ & $0.9522 \pm 0.0478$ & $\underline{0.1626} \pm 0.0200$ \\
S5   & 10 & $0.4955 \pm 0.0183$ & $0.8153 \pm 0.0924$ & $0.3459 \pm 0.0518$ & 10 & $0.5133 \pm 0.0227$ & $0.8194 \pm 0.0606$ & $0.3697 \pm 0.0353$ \\
S6.1 & 9  & $0.4803 \pm 0.0274$ & $0.8153 \pm 0.0916$ & $0.3383 \pm 0.0537$ & 8  & $0.5010 \pm 0.0227$ & $0.8194 \pm 0.0628$ & $0.3635 \pm 0.0382$ \\
S6.2 & 12 & $0.6059 \pm 0.0256$ & $0.8153 \pm 0.0943$ & $0.4011 \pm 0.0562$ & 12 & $0.5152 \pm 0.0240$ & $0.7845 \pm 0.0725$ & $0.3955 \pm 0.0415$ \\
S7.1 & 5  & $0.4699 \pm 0.0198$ & $\underline{0.9846} \pm 0.0231$ & $0.2431 \pm 0.0162$ & 7  & $0.4867 \pm 0.0267$ & $0.7878 \pm 0.0732$ & $0.3581 \pm 0.0414$ \\
S7.2 & 6  & $0.4699 \pm 0.0193$ & $0.9624 \pm 0.0453$ & $0.2551 \pm 0.0286$ & 6  & $0.4867 \pm 0.0272$ & $0.8948 \pm 0.0547$ & $0.2885 \pm 0.0297$ \\
Baseline & 11 & $0.4991 \pm 0.0157$ & $0.8153 \pm 0.0921$ & $0.3477 \pm 0.0530$ & 11 & $0.5199 \pm 0.0223$ & $0.8194 \pm 0.0612$ & $0.3730 \pm 0.0350$ \\
\bottomrule
\end{tabular}
% \vspace{-1 em}
\end{table*}

\paragraph{\textbf{Visual interaction cues and multimodal fusion:}}
S7 is the clearest example of a third strategy ( conversation clustering from visual interaction cues). Instead of relying purely on temporal overlap or semantic content, it estimates gaze directions with UniGaze\cite{Unigaze}, computes mutual gaze scores between speakers, and uses these signals as evidence of conversational linkage. To improve robustness, the gaze-based similarity is fused with the baseline overlap-based score before AHC. This design is particularly interesting because it operates on interaction structure more directly than transcript-based approaches. However, this approach is sensitive to errors in gaze estimation, as head-pose variation, occlusion, and weak mutual gaze can weaken or distort interaction signals even among speakers in the same conversation. S7 therefore illustrates both the promise and the limitations of going beyond speech-only interaction cues.

\vspace{-0.5em}
\subsection{Cross-Cutting Design Patterns}

First, nearly all systems remain modular. Even when they use stronger backbones, explicit extraction, or LLM-based reasoning, transcription and clustering are still solved as separate stages connected by intermediate outputs such as timestamps or transcripts. End-to-end modeling of ``who speaks what, when, and to whom'' remains largely absent.

Second, systems differ not only by modality, but by \textit{where} the modality is used. Visual information supports segmentation (most systems except S4), extraction (S1, S5, S6), recognition (all systems), or clustering (S7). Likewise, language models may appear as post-processors for transcript correction (S1, S2), or as LLM-based semantic clustering components that infer speaker groupings from transcripts (S1, S2, S4).

\vspace{-1em}
\subsection{Training Data and Supervision}

Beyond architectural design, the systems differ in their use of training data and development strategies. As shown in Table~\ref{tab:dataset_usage_trained_only}, most approaches combine large-scale external audio-visual corpora (e.g., LRS2, LRS3, VoxCeleb2, AVSpeech, AVYT) with the MCoRec dataset, but adopt distinct strategies for leveraging them. S1, S2, S4, and S5 follow a multi-stage training paradigm, where AVSR models are first pretrained on large-scale or simulated mixtures and then fine-tuned on MCoRec to adapt to real overlapping conversational conditions. S1 further extends this strategy to additional components, jointly training ASD and AVTSE, while S5 incorporates AVTSE as an auxiliary task during training. In contrast, S6 relies on a pseudo-supervised approach, using enhanced close-talk signals derived from MCoRec as supervision for training AVTSE. S3 and S7 take a different direction, avoiding model retraining and instead tuning hyperparameters directly on MCoRec.

\vspace{-0.5em}
\section{Results}

Table~\ref{tab:dev_eval_results} reveals several consistent trends across systems. First, JointError is largely driven by WER once clustering F1 exceeds approximately 0.95. For example, S1 and S4 achieve similar clustering performance on the evaluation set (0.9572 vs. 0.9522), yet their JointError closely follows their WER (0.3018 vs. 0.3034), indicating that transcription dominates in this range. While clustering becomes less influential for top systems at this level, it remains a differentiating factor for lower-performing systems where F1 drops below 0.90 (e.g., S7.1: 0.7878 vs. S7.2: 0.8948, with corresponding JointError 0.3581 vs. 0.2885). Second, systems leveraging LLM-based semantic clustering achieve consistently high F1 scores (e.g., S1, S2, S4 all above 0.95), substantially outperforming the baseline clustering (0.8194). Third, stronger AVSR modeling, whether through highly optimized pipelines (S1) or long-context architectures (S4), consistently outperforms approaches relying on explicit target-speaker extraction (S6), with large WER gaps on evaluation (e.g., S1: 0.3018 vs. S6.1: 0.5010), indicating that improving robustness within the recognition model is more effective than front-end separation. Fourth, LLM-based post-correction provides limited gains under high-WER conditions, as seen in the small differences between S2 variants (e.g., S2.1: 0.4525 vs. S2.2: 0.4451 WER). Fifth, the multimodal cue-based approache (S7) performs well on the development set (F1 up to 0.9846) but degrades on evaluation (as low as 0.7878), suggesting weaker generalization compared to semantic clustering approaches. Finally, reducing reliance on strict short-segment ASD, either through stronger detection (S1) or longer-context inference (S4), appears beneficial, although ASD remains a core component in all systems. Overall, these results indicate that improvements in AVSR modeling are the main driver of performance once clustering performance is high, while clustering and post-processing play a secondary but still necessary role.

\begin{table*}[htbp]
\centering
\caption{Performance on the evaluation set by number of speakers.}
\label{tab:eval_by_num_speakers}
\renewcommand{\arraystretch}{1.0}
\setlength{\tabcolsep}{5pt}
\small
\resizebox{\textwidth}{!}{%
\begin{tabular}{@{}l *{14}{c}@{}}
\toprule
\multirow{2}{*}{\textbf{System}} &
\multicolumn{2}{c}{\textbf{2 spk}} &
\multicolumn{2}{c}{\textbf{3 spk}} &
\multicolumn{2}{c}{\textbf{4 spk}} &
\multicolumn{2}{c}{\textbf{5 spk}} &
\multicolumn{2}{c}{\textbf{6 spk}} &
\multicolumn{2}{c}{\textbf{7 spk}} &
\multicolumn{2}{c}{\textbf{8 spk}} \\
\cmidrule(lr){2-3} \cmidrule(lr){4-5} \cmidrule(lr){6-7}
\cmidrule(lr){8-9} \cmidrule(lr){10-11} \cmidrule(lr){12-13} \cmidrule(l){14-15}
& \textbf{WER $\downarrow$} & \textbf{F1 $\uparrow$}
& \textbf{WER $\downarrow$} & \textbf{F1 $\uparrow$}
& \textbf{WER $\downarrow$} & \textbf{F1 $\uparrow$}
& \textbf{WER $\downarrow$} & \textbf{F1 $\uparrow$}
& \textbf{WER $\downarrow$} & \textbf{F1 $\uparrow$}
& \textbf{WER $\downarrow$} & \textbf{F1 $\uparrow$}
& \textbf{WER $\downarrow$} & \textbf{F1 $\uparrow$} \\
\midrule
S1 &
\underline{0.219} & \textbf{1.00} &
\underline{0.295} & \textbf{1.00} &
\underline{0.249} & \textbf{1.00} &
\textbf{0.354} & \textbf{1.00} &
\underline{0.282} & \textbf{0.97} &
\textbf{0.509} & \textbf{1.00} &
\textbf{0.436} & \textbf{1.00} \\
S2.1 &
0.441 & \textbf{1.00} &
0.509 & \textbf{1.00} &
0.398 & \textbf{1.00} &
0.502 & \textbf{1.00} &
0.424 & \underline{0.96} &
0.618 & \textbf{1.00} &
0.582 & \textbf{1.00} \\
S2.2 &
0.434 & \textbf{1.00} &
0.491 & \textbf{1.00} &
0.390 & \textbf{1.00} &
\underline{0.493} & \textbf{1.00} &
0.417 & \underline{0.96} &
0.613 & \textbf{1.00} &
0.574 & \textbf{1.00} \\
S2.3 &
0.414 & \underline{0.92} &
0.519 & \textbf{1.00} &
0.427 & \textbf{1.00} &
0.533 & \textbf{1.00} &
0.472 & \underline{0.96} &
0.705 & \underline{0.80} &
0.713 & 0.85 \\
S3 &
0.548 & \underline{0.92} &
0.559 & \textbf{1.00} &
0.555 & 0.87 &
0.576 & 0.79 &
0.480 & 0.79 &
0.646 & \textbf{1.00} &
0.627 & 0.70 \\
S4 &
\textbf{0.214} & \textbf{1.00} &
\textbf{0.290} & \textbf{1.00} &
\textbf{0.248} & \textbf{1.00} &
\textbf{0.354} & \textbf{1.00} &
\textbf{0.281} & 0.95 &
\underline{0.514} & \textbf{1.00} &
\underline{0.468} & \textbf{1.00} \\
S5 &
0.543 & \underline{0.92} &
0.573 & \textbf{1.00} &
0.465 & 0.92 &
0.575 & 0.82 &
0.476 & 0.78 &
0.657 & \underline{0.80} &
0.621 & 0.56 \\
S6.1 &
0.488 & \underline{0.92} &
0.524 & \textbf{1.00} &
0.448 & 0.92 &
0.547 & 0.82 &
0.472 & 0.78 &
0.685 & \underline{0.80} &
0.637 & 0.56 \\
S6.2 &
0.498 & \underline{0.92} &
0.552 & \textbf{1.00} &
0.467 & 0.75 &
0.564 & \underline{0.88} &
0.482 & 0.79 &
0.681 & 0.40 &
0.668 & 0.36 \\
S7.1 &
0.453 & 0.83 &
0.497 & \underline{0.50} &
0.428 & 0.90 &
0.555 & 0.68 &
0.453 & 0.75 &
0.635 & \textbf{1.00} &
0.643 & \underline{0.90} \\
S7.2 &
0.453 & \underline{0.92} &
0.497 & \underline{0.50} &
0.428 & \underline{0.95} &
0.555 & 0.87 &
0.453 & 0.92 &
0.635 & 0.62 &
0.643 & \textbf{1.00} \\
Baseline &
0.544 & \underline{0.92} &
0.557 & \textbf{1.00} &
0.477 & 0.92 &
0.576 & 0.82 &
0.483 & 0.78 &
0.636 & \underline{0.80} &
0.641 & 0.56 \\
\bottomrule
\end{tabular}%
}
\vspace{-1 em}
\end{table*}

Tables~\ref{tab:eval_by_num_speakers} and~\ref{tab:eval_by_num_conv} provide complementary views of system behavior under increasing conversational complexity. WER generally increases as the number of speakers grows across all systems, reflecting the increasing impact of overlap and interference. In contrast, varying the number of conversations has a limited and less consistent effect on WER, which remains relatively stable across 2--3 conversations (e.g., S1: 0.294 → 0.304). Comparing these two trends suggests that transcription difficulty is more sensitive to speaker interference than to the number of conversational groups. Clustering performance shows the opposite pattern. F1 remains near-saturated for top systems in simpler settings (fewer speakers or a single conversation), but degrades as conversational structure becomes more complex, particularly in multi-conversation scenarios. Systems with weaker AVSR quality or non-semantic clustering approaches (e.g., S3, S5, S6, S7) show clear drops in F1 as the number of conversations increases (e.g., S3: 1.00 → 0.77 → 0.83), while LLM-based approaches (S1, S2, S4) maintain consistently high clustering performance. Interestingly, S2 consistently achieves high clustering performance despite high WER. Overall, increasing the number of speakers primarily affects WER, while increasing the number of conversations mainly impacts clustering performance. Strong systems (S1, S4) remain stable across both conditions, whereas systems with weaker AVSR or non-semantic clustering degrade more noticeably, particularly in multi-conversation settings.

\begin{table}
\caption{Benchmarking evaluation set by number of\\ conversations.}
% \vspace{-0.5 em}
\label{tab:eval_by_num_conv}
\begin{tabular}{@{}lcccccc@{}}
\toprule
\multirow{2}{*}{\textbf{System}} &
\multicolumn{2}{c}{\textbf{1 conv}} &
\multicolumn{2}{c}{\textbf{2 conv}} &
\multicolumn{2}{c}{\textbf{3 conv}} \\
\cmidrule(lr){2-3} \cmidrule(lr){4-5} \cmidrule(l){6-7}
& \textbf{WER $\downarrow$} & \textbf{F1 $\uparrow$}
& \textbf{WER $\downarrow$} & \textbf{F1 $\uparrow$}
& \textbf{WER $\downarrow$} & \textbf{F1 $\uparrow$} \\
\midrule
S1       & \underline{0.225} & \textbf{1.00} & \underline{0.294} & \textbf{1.00} & \textbf{0.304} & \textbf{0.98} \\
S2.1     & 0.456             & \textbf{1.00} & 0.457             & \underline{0.99} & 0.433           & \textbf{0.98} \\
S2.2     & 0.453             & \textbf{1.00} & 0.447             & \underline{0.99} & 0.427           & \textbf{0.98} \\
S2.3     & 0.425             & \textbf{1.00} & 0.473             & 0.97             & 0.505           & 0.94 \\
S3       & 0.572             & \textbf{1.00} & 0.562             & 0.77             & 0.490           & 0.83 \\
S4       & \textbf{0.215}    & \textbf{1.00} & \textbf{0.290}    & \textbf{1.00}    & \underline{0.312} & \underline{0.96} \\
S5       & 0.569             & \textbf{1.00} & 0.522             & 0.82             & 0.485           & 0.79 \\
S6.1     & 0.502             & \textbf{1.00} & 0.501             & 0.82             & 0.486           & 0.79 \\
S6.2     & 0.514             & \textbf{1.00} & 0.513             & 0.76             & 0.502           & 0.77 \\
S7.1     & 0.466             & \textbf{1.00} & 0.490             & 0.71             & 0.470           & 0.80 \\
S7.2     & 0.466             & \textbf{1.00} & 0.490             & 0.86             & 0.470           & 0.93 \\
Baseline & 0.565             & \textbf{1.00} & 0.529             & 0.82             & 0.493           & 0.79 \\
\bottomrule
\end{tabular}
\end{table}

Table~\ref{tab:eval_by_activity} analyzes system performance under different levels of speaking activity. The speaking activity ratio is defined as the proportion of time a speaker is active relative to the session duration. Speakers are categorized as low ($< 0.45$), mid ($0.45$--$0.60$), or high ($\geq 0.60$) activity, with 24.1\%, 26.9\%, and 49.1\% of speakers in each group, respectively. A consistent trend across systems is that WER decreases as speaking activity increases (e.g., S1: 0.371 → 0.271; S2.2: 0.554 → 0.386), indicating that low-activity speakers with more fragmented and sparse speech are more challenging for transcription. Clustering performance does not show a consistent trend with speaking activity. While some systems improve with more speech (e.g., S3, S7), others degrade (e.g., S5, S6), indicating that higher activity introduces not only more evidence but also more overlap and cross-speaker interference. In contrast, top systems (S1, S4) maintain both low WER and high F1 across all activity levels, suggesting greater robustness to fragmented conversational structure. Overall, these results show that beyond speaker density and conversational structure, temporal continuity of speech is another key factor influencing both recognition and clustering performance in the MCoRec setting.

\begin{table}
\caption{Benchmarking the evaluation set by speaking \\activity, where Low denotes activity $< 0.45$, Mid denotes \\activity $0.45$--$0.60$, and High denotes activity $\geq 0.60$.}
\label{tab:eval_by_activity}
\begin{tabular}{@{}lcccccc@{}}
\toprule
\multirow{2}{*}{\textbf{System}} &
\multicolumn{2}{c}{\textbf{Low}} &
\multicolumn{2}{c}{\textbf{Mid}} &
\multicolumn{2}{c}{\textbf{High}} \\
\cmidrule(lr){2-3} \cmidrule(lr){4-5} \cmidrule(l){6-7}
& \textbf{WER $\downarrow$} & \textbf{F1 $\uparrow$}
& \textbf{WER $\downarrow$} & \textbf{F1 $\uparrow$}
& \textbf{WER $\downarrow$} & \textbf{F1 $\uparrow$} \\
\midrule
S1       & \underline{0.371} & \underline{0.98} & \underline{0.296} & \textbf{0.99} & \textbf{0.271} & \textbf{0.98} \\
S2.1     & 0.563             & 0.97 & 0.462             & \textbf{0.99} & 0.393             & \textbf{0.98} \\
S2.2     & 0.554             & 0.97 & 0.456             & \textbf{0.99} & 0.386             & \textbf{0.98} \\
S2.3     & 0.596             & 0.97             & 0.510             & 0.95           & 0.438             & 0.96 \\
S3       & 0.680             & 0.73             & 0.544             & 0.79           & 0.450             & 0.85 \\
S4       & \textbf{0.364}    & \textbf{0.99}    & \textbf{0.292}    & \textbf{0.99}  & \underline{0.280} & \underline{0.97} \\
S5       & 0.637             & 0.80             & 0.542             & 0.78           & 0.437             & 0.75 \\
S6.1     & 0.593             & 0.80             & 0.503             & 0.75           & 0.445             & 0.64 \\
S6.2     & 0.601             & 0.80             & 0.516             & 0.78           & 0.443             & 0.75 \\
S7.1     & 0.592             & 0.74             & 0.502             & 0.74           & 0.427             & 0.80 \\
S7.2     & 0.592             & 0.90             & 0.502             & 0.91           & 0.427             & 0.91 \\
Baseline & 0.634             & 0.80             & 0.543             & 0.78           & 0.451             & 0.75 \\
\bottomrule
\end{tabular}
\end{table}

% \section{Discussion}

% Our analysis shows that systems do not fail only because of overlap. In session\_12, session\_14, session\_147, and session\_151 each evaluated speaker spends about 90--95\% of their speaking time overlapping with at least three other speakers, and all systems perform poorly (WER $\approx$ 0.4--0.5). This suggests a global failure case under extreme overlap. However, in session\_123, session\_125, session\_13, all speakers also have very high overlap (mean overlap $\sim$0.97--1.0), but performance varies a lot across speakers (e.g., WER from $\sim$0.095 to $\sim$0.457). This shows that overlap alone does not explain the difference between easy and hard speakers. Instead, we observe that high-error speakers have more cross-speaker confusion, with about 31--53\% of substitution words matching other speakers, compared to $\sim$14\% for the easier speaker. In addition, the easier speaker has more continuous speech, with shorter gaps and a larger share of reference words. These results suggest that systems fail either when overlap is consistently high for all speakers, or when the target speech is fragmented or easily confused with competing speakers.

To better understand system failures beyond simple overlap effects, we analyze the top AVSR systems (S1 and S4) at the session. In session\_12, session\_14, session\_147, and session\_151, all speakers exhibit extreme overlap (90--95\% of speaking time overlapping with at least three others), and all systems perform poorly (WER $\approx$ 0.45--0.6), indicating a global failure case under uniformly high interference. However, in session\_123, session\_125, and session\_13, speakers also experience similarly high overlap (mean $\sim$0.97--1.0), yet performance varies widely across speakers (e.g., WER $\sim$0.095 to $\sim$0.457). This shows that overlap alone does not explain performance differences. Instead, high-error speakers exhibit stronger cross-speaker confusion, with 31--53\% of substitution words matching other speakers, compared to $\sim$14\% for easier speakers. They also tend to have more fragmented speech, with shorter segments and fewer reference words. Overall, systems fail either under uniformly high interference across all speakers, or when the target speech is fragmented and easily confused with competing speakers.

% \vspace{-1em}
\section{Conclusion}
This paper analyzed the CHiME-9 MCoRec submissions from a system-level perspective, focusing on how different design choices address the core challenge of recognizing and organizing speech in multi-party, overlapping conversations. While the task is often framed as a severe overlap problem, our results show that overlap alone does not explain performance differences. Instead, the main difficulty lies in correctly associating speech with the target speaker and maintaining consistency across fragmented conversational structure. Across systems, we observe three distinct strategies: (i) strengthening audio-visual representations for target-speaker recognition, (ii) explicitly handling interference through extraction or auxiliary objectives, and (iii) leveraging higher-level reasoning, especially with large language models, to resolve conversation structure. These strategies address complementary failure modes, and no single approach is sufficient. Systems integrating multiple such components achieve up to 57\% relative reduction in the joint error rate. This suggests that the cocktail-party problem is not dominated by a single factor such as overlap, but by multiple interacting sources of ambiguity that require coordinated modeling across components.

% \section{Acknowledgment}
% \vspace{-0.5em}
\begin{acks}
The authors gratefully acknowledge support from the EU’s Horizon research \& innovation programme (101135798 – Meetween; 101213369 – DVPS). We thank Interactive-AI LLC for providing the database for research purposes.
\end{acks}
\bibliographystyle{ACM-Reference-Format}
\bibliography{sample-base}

\end{document}